\pdfoutput=1
\documentclass[11pt]{article}

\usepackage{ACL2023}

\usepackage{times}
\usepackage{latexsym}

\usepackage[T1]{fontenc}
\usepackage[utf8]{inputenc}
\usepackage{caption}
\usepackage{subcaption}
\usepackage{adjustbox}
\usepackage{float}
\restylefloat{table}
\usepackage{wrapfig,lipsum,booktabs}
\usepackage{array,multirow,graphicx}
\usepackage{array}
\usepackage{arydshln}

\usepackage{appendix}
\usepackage{microtype}

\usepackage{inconsolata}

\title{UBC-DLNLP at SemEval-2023 Task 12: \\Impact of Transfer Learning on African Sentiment Analysis}

\author{First Author \\
  Affiliation / Address line 1 \\
  Affiliation / Address line 2 \\
  Affiliation / Address line 3 \\
  \texttt{email@domain} \\\And
  Second Author \\
  Affiliation / Address line 1 \\
  Affiliation / Address line 2 \\
  Affiliation / Address line 3 \\
  \texttt{email@domain} \\}

\author{\normalsize Gagan Bhatia$^{1,\star}$ ~ Ife Adebara$^{1,\star}$ ~ AbdelRahim Elmadany$^{1}$ ~ Muhammad Abdul-Mageed$^{1,2}$ \\
\normalsize $^{1}$Deep Learning \& Natural Language Processing Group,
  The University of British Columbia\\\normalsize  $^{2}$Department of Natural Language Processing \& Department of Machine Learning, MBZUAI\\ %
  \texttt{\normalsize \{gagan30@student.,ife.adebara@,a.elmadany@,muhammad.mageed@\}ubc.ca}}
  
\begin{document}
\maketitle
\begin{abstract}
We describe our contribution to the SemEVAl 2023 AfriSenti-SemEval shared task, where we tackle the task of sentiment analysis in $14$ different African languages. We develop both monolingual and multilingual models under a full supervised setting (subtasks A and B). We also develop models for the zero-shot setting (subtask C). Our approach involves experimenting with transfer learning using six language models, including further pretraining of some of these models as well as a final finetuning stage. Our best performing models achieve an \textit{F\textsubscript{1}}-score of $70.36$ on development data and an \textit{F\textsubscript{1}}-score of $66.13$ on test data. Unsurprisingly, our results demonstrate the effectiveness of transfer learning and finetuning techniques for sentiment analysis across multiple languages. Our approach can be applied to other sentiment analysis tasks in different languages and domains.
\end{abstract}

\section{Introduction}

Sentiment Analysis, also referred to as opinion mining, is a Natural Language Processing (NLP) technique that aims to identify, extract, and evaluate opinions, attitudes, perceptions, and sentiments towards topics, products, services, and individuals from textual data \cite{BIRJALI2021107134}. With the increased accessibility of the internet, people are increasingly sharing their opinions on various platforms, such as forums, blogs, wikis, websites, and social media pages. Consequently, there is a need for the automatic extraction of sentiments to gain valuable insights into user perception, popular opinion, and trends  \cite{GEORGIADOU2020102048, tinoco-2018}. 

Despite the increasing popularity of sentiment analysis, its application in low-resource African languages is still under-explored \cite{shode_africanlp, diallo2021bambara}. This is because many African languages have limited digital resources, such as annotated data and lexical resources, which can hinder the development and evaluation of sentiment analysis models. So far, only a handful of African languages have few datasets for sentiment analysis \cite{abubakar_2021, ogbuju-onyesolu-2019-development, muhammadSemEval2023,  afrisenti_2023, oyewusi_2020, muhammad-etal-2022-naijasenti}. Furthermore, African languages often exhibit complex morphology, syntax, semantics, stylistics, pragmatic and orthographic conventions including the use of diacritics, and code-mixing that can make it difficult to accurately identify and extract sentiment from text data \cite{muhammadSemEval2023, afrisenti_2023, orimaye_2012}. For instance, for some African languages, a single change in tone assignment can change the sentiment of a text \cite{adebara-abdul-mageed-2022-towards}. 
\begin{figure}[t]
  \centering
  \includegraphics[width=\columnwidth]{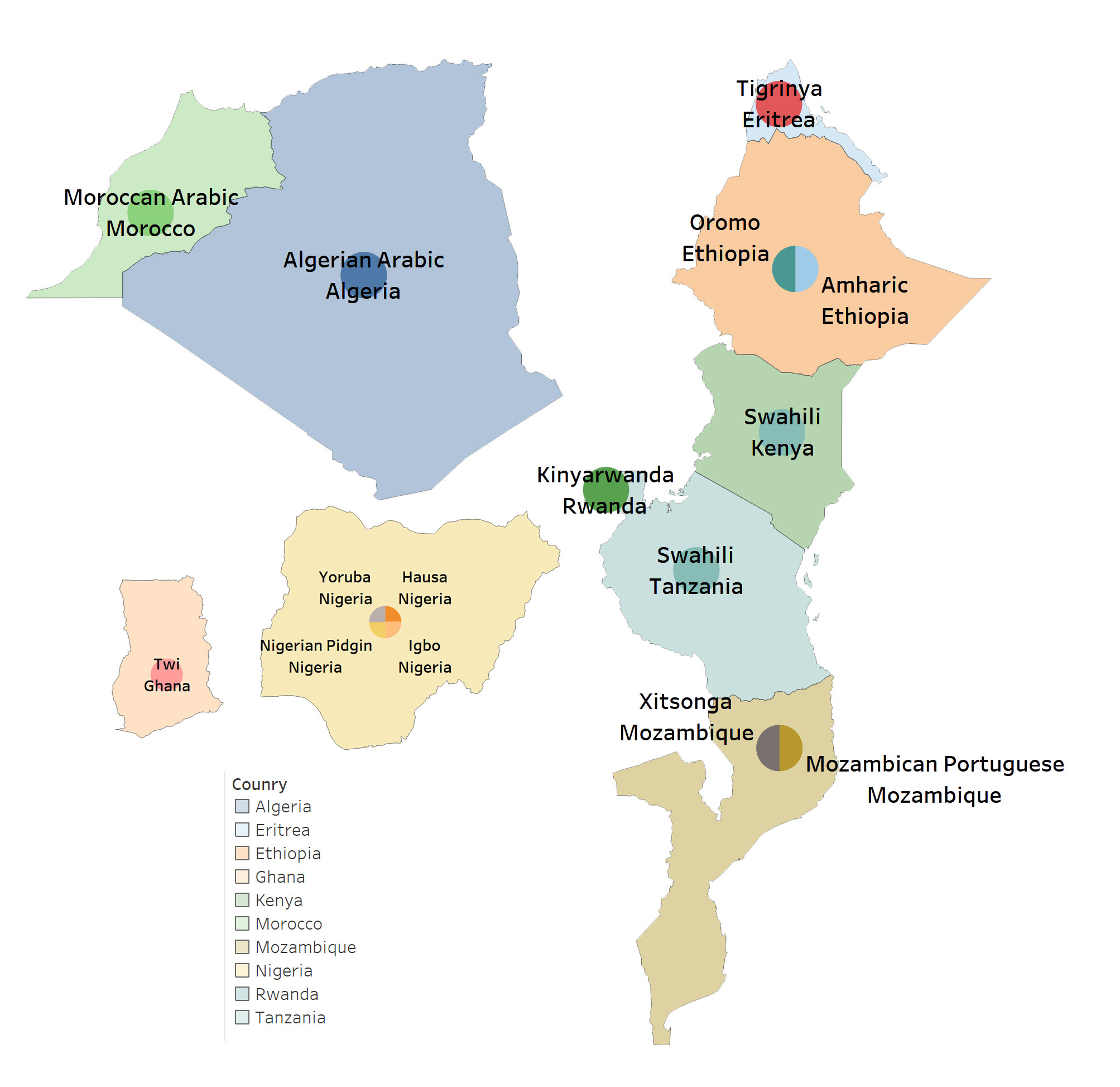}
\caption{\small A map showing the countries where each language in the shared task is spoken in Africa. }
\label{fig:countries} 
\end{figure}
% For instance, sentiment information in product reviews can influence potential customers' perception and provide product owners with crucial information on their customers' needs and behavioral patterns. By analyzing and comprehending customer reviews' sentiments, businesses can engage better with their customers, enhance their products and services, and ultimately improve customer satisfaction and loyalty. Furthermore, sentiment analysis has been used to predict outcomes of elections \cite{tinoco-2018} and other crucial decisions \cite{GEORGIADOU2020102048}.

 In this task, we conduct sentiment analysis on $14$ African languages including Algerian Arabic, Amharic, Hausa, Igbo, Kinyarwanda, Moroccan Arabic, Mozambican Portuguese, Nigerian Pidgin, Oromo, Swahili, Tigrinya,  Twi, and Yoruba. The sentiment analysis data used for this shared task is the largest and most multilingual dataset for sentiment analysis for African languages to date \cite{muhammadSemEval2023, afrisenti_2023}.

Our contribution is as follows:
\begin{enumerate}
    \item We show the utility of finetuning six language models for sentiment analysis on $14$ African languages.
    \item We show the utility of further pretraining two language models for sentiment analysis on $14$ African languages. 
    \item We show the performance of our models in zero-shot settings. 
\end{enumerate}

The rest of this paper is organized as follows: We discuss existing literature in Section \ref{sec:litreview}, and provide background information in Section \ref{sec:approach}. Section \ref{sec:system_overview} has details about the models we develop. In Section \ref{sec:experiment} we describe each experiment performed and show the results on Dev. and Test sets in Section \ref{sec:results}. We conclude in Section \ref{sec:conc}.

\section{Literature Review}\label{sec:litreview}
\subsection{Sentiment Analysis}

Sentiment analysis can be conceptualized as a text classification problem, where the sentiment of the text is classified into one of three categories: negative, neutral, or positive. Different levels of sentiment analysis include document level \cite{Behdenna2016159}, sentence level, and aspect level \cite{Do2019272, xue-li-2018-aspect}. Document level analysis focuses on the overall sentiment of a text, whereas sentence level analysis evaluates sentiment on a more fine-grained level. Aspect level analysis focuses on specific features in the text.

The methods for sentiment analysis have evolved rapidly, from rule-based approaches \cite{turney-2002-thumbs} to machine learning, deep learning, and hybrid methods \cite{akhtar-etal-2016-hybrid}. Rule-based methods rely on identifying polarity items \cite{wilson-etal-2005-recognizing, medhaffar-etal-2017-sentiment}, punctuation, and other linguistic features to determine sentiment. Although these methods are easy to interpret and implement, developing rules can be tedious, expensive, and lack scalability. Machine learning approaches like support vector machines and Naive Bayes learn from labeled data to predict sentiment in new, unlabeled text. Deep learning methods, including convolutional neural networks  \cite{dos-santos-gatti-2014-deep, xue-li-2018-aspect}, transformers, and transfer learning approaches \cite{baert-etal-2020-arabizi, sun-etal-2019-utilizing, hosseini-asl-etal-2022-generative}, have achieved state-of-the-art performance in sentiment analysis. In hybrid methods \cite{akhtar-etal-2016-hybrid}, two or more of the aforementioned methods are combined for sentiment analysis. Hybrid methods and Transfer learning methods are able to achieve high accuracy in low resource scenarios. 

\subsection{Transfer Learning}
Transfer learning \cite{JMLR:v21:20-074, he2022towards, ruder-etal-2019-transfer, sebastianruder_2022} is an integral part of modern NLP systems. Transfer learning attempts to transfer knowledge from other sources to benefit a current task; based on the premise that previous knowledge may improve solutions for a current task \cite{5288526}. It allows the domains, tasks, and distributions used in training and testing to be different, enabling a new task to leverage previously acquired domain knowledge. Potential benefits include faster learning, better generalization, and a more robust system.  It has significantly improved state of the art in natural language generation (NLG) and natural language understanding (NLU) tasks of which Sentiment Analysis is one. Transfer learning, through the use of large transformer models have enabled the use of low-resource languages through finetuned on various NLP tasks. 

In monolingual settings, transfer learning involves using pre-trained models on data in one language while multilingual transfer learning involves using pre-trained models on large datasets in multiple languages \cite{DBLP:journals/corr/abs-2108-10640}. The multilingual transfer learning approach takes advantage of the fact that many languages share similar structures and patterns, which can be leveraged to improve performance in low resource languages \cite{ruder-etal-2019-transfer, sebastianruder_2022}. In this work, we experiment with language models that have representations of some African languages to transfer representations for our sentiment analysis task. We also experiment with monolingual and multilingual settings. In addition, we perform two experiments in zero-shot settings. 

% The growing use of the Internet have made the web become the universal and the most important source of information. Millions of people express their opinions, and sentiments in forums, blogs, wikis, social networks, and other web resources [14], [15], [16]. Those opinions and sentiments are very relevant to our daily lives, and hence there is a need to analyze this user-generated data in order to automatically monitor the public opinion and assist decision-making [6], [14]. For example, Twitter posts have been used to predict election results [17]. Sentiment are relevant to our daily lives, necessitating the need for their automatic extraction from text. 

\subsection{African NLP}
Africa is home to over 2,000 Indigenous languages, which represents about one-third of all languages spoken globally \cite{ethnologue}. Despite this, most of these languages have not received much attention in the field of Natural Language Processing (NLP). Unfortunately, the majority of NLP research has focused on higher-resource languages, which are typologically distinct from Indigenous African languages. The methods used to develop NLP technologies for these languages have been Western-centric, making them challenging to apply directly to African languages \cite{adebara-abdul-mageed-2022-towards}. Additionally, existing NLP technologies function within the context of Western values and beliefs, which poses unique challenges when these technologies are applied within African communities.

To address this language bias problem, an Afrocentric approach to technology development is crucial for African languages. Such an approach would entail developing technologies that meet the needs of local African communities \cite{adebara-abdul-mageed-2022-towards}. Several NLP It would involve not only deciding what technologies to build but also determining how to build, evaluate, and deploy them \cite{adebara-etal-2022-afrolid}. By adopting an Afrocentric approach, NLP researchers and practitioners can help to bridge the digital divide and ensure that language technologies are accessible to African communities.

\section{Approach}\label{sec:approach}
We perform sentiment analysis on three different subtasks with $14$ languages spoken across Africa. The languages are quite diverse belonging to four different language families and written in different scripts including Arabic, Ethiopic, and Latin scripts. We provide details about the languages and the datasets. 

% Sentiment analysis deals with the classification and detection of sentiment in texts. Most sentiment analysis models focus on English or other high-resource languages, While lesser-spoken languages are often lesser represented. There has been a growing interest in sentiment analysis which applies to many domains, including public health, commerce/business, art and literature, social sciences, neuroscience, and psychology. 
 
% In this shared task, we were given datasets in 17 different languages. In this task, we have 17 African languages, Hausa, Yoruba, Igbo, Nigerian Pidgin from Nigeria, Amharic, Tigrinya, and Oromo from Ethiopia, Swahili from Kenya and Tanzania, Algerian Arabic dialect from Algeria, Kinyarwanda from Rwanda, Twi from Ghana, Mozambique Portuguese from Mozambique  and Moroccan Arabic/Darija from Morocco. 

\subsection{Datasets}
This study utilizes Twitter datasets provided for the SemEVAl 2023 AfriSenti-SemEval shared task \cite{muhammad2023afrisenti}. The dataset comprises three subtasks, each with a different focus on sentiment analysis. \textbf{Subtask A} consists of monolingual datasets for $12$ different languages, each labeled as positive, negative, or neutral. \textbf{Subtask B} involves a multilingual sentiment analysis system, with multilingual data for the $12$ languages in Task A. \textbf{Subtask C} provides unlabeled data for two African languages (Tigrinya and Oromo), and participants are expected to develop a zero-shot model for sentiment analysis in these languages. The dataset statistics for each language are presented in detail in Table \ref{tab:stats}. The use of Twitter datasets enables the evaluation of sentiment analysis models on real-world data, providing insights into the effectiveness of different approaches for sentiment analysis in a multilingual context. We provide details of each language in Table \ref{tab:langsinfo} and Section \ref{app:apx}. For preprocessing, we remove all URLs and tokenize with wordpiece.
\begin{table}[!h]
\small
\centering
\begin{tabular}{llrrr}
\toprule
\textbf{Subsets} & \textbf{Subtask} & \textbf{Train} & \textbf{Dev} & \textbf{Test} \\
\midrule
am           & A & 8,978                 & 1,498  & 2,000  \\
dz           & A & 2,479                 & 415   & 959   \\
ha           & A & 19,526                & 2,678  & 5,304  \\
ig           & A & 13,874                & 1,842  & 3,683  \\
kr           & A & 4,956                 & 828   & 1,027  \\
ma           & A & 8,013                 & 495   & 2,962  \\
sw           & A & 2,716                 & 454   & 749   \\
pcm          & A & 7,683                 & 1,282  & 4,155  \\
pt           & A & 4,597                 & 768   & 3,663  \\
ts           & A & 1,210                 & 204   & 255   \\
twi          & A & 4,257                 & 389   & 950   \\
yo           & A & 12,702                & 2,091  & 4,516  \\ 
\midrule 
multilingual & B & 90,991                & 13,654 & 30,212 \\ 
\midrule 
or           & C &  ------& 397   & 2,097  \\
tg           & C &  ------& 399   & 2,001 \\

\bottomrule
\end{tabular}\caption{Statistics of data for each language across the three tasks. \textbf{am}: Amharic, \textbf{dz}: Algerian Arabic, \textbf{ha}: Hausa, \textbf{ig}: Igbo, \textbf{kr}: Kinyarwanda , \textbf{ma}: Darija, \textbf{sw}: Swahili, \textbf{pcm}: Nigerian Pidgin, \textbf{pt}: Mozambican Portuguese, \textbf{ts}: Xitsonga (Mozambique Dialect), \textbf{twi}: Twi, \textbf{yo}: Yoruba, \textbf{or}: Oromo, \textbf{tg}: Tigrinya.}
\label{tab:stats} 
\end{table}

\subsection{Code and Script Switching}
We found examples of code-switching and script switching in the data used for training. Moroccan Arabic data for instance had both Arabic and Latin scripts examples. We also found code-mixing with English in the Hausa, Igbo, Twi, Swahili, and Yoruba and code-mixing with French in the Algerian Arabic examples.

%Each dataset was labelled with positive, negative, and neutral labels.  In each dataset we have all the training data labeled as with their polarity of the tweet as positive, negative and neutral. In the training data if we have both positive and negative sentiment the sentiment with higher polarity of the sentiment. 
\section{System Overview}\label{sec:system_overview}

In order to identify the best-performing model for our datasets, we first finetuned 6 LMs on the data from sub tasks A and B. Specifically, we finetuned mBERT, XLM-R, Afro-XLMR, AfriBERTa, AfriTEVA, and Serengeti. We also further pretrained Afro-XLMR and Serengeti. We refer to the pre-trained models as Afro-XLMR-LM and Serengeti-LM, respectively. We provide further details for each of the LMS in what follows. 
% We also summarize the information about the models in Table \ref{tab:encoder-only} and the African languages represented in each model in Table \ref{tab:resources}.

\subsection{Models} 
\subsubsection{XLM-R}

XLM-R \cite{DBLP:journals/corr/abs-1911-02116} is an encoder-only model based on RoBERTa. It was pretrained on a corpus of 100 languages, of which only 8 were African. Namely  Afrikaans, Amharic, Hausa, Oromo, Somali, Swahili, Xhosa, out of which Oromo, Hausa and Swahili are part of the shared task.  We use finetune both base and large models. 

\subsubsection{mBERT}

mBERT \cite{DBLP:journals/corr/abs-1810-04805} is a multilingual variant of BERT pretrained on $104$ languages. Out of these $104$ languages only $4$ languages are African out of which Swahili and Yoruba are part of this shared task. mBERT was pre-trained using masked language modeling (MLM) and next-sentence prediction task. We finetune the base model. 

\subsubsection{Afro-XLMR}

Afro-XLM-R \cite{alabi-etal-2022-adapting}  uses language adaptation on the $17$ most-resourced African languages and three other high-resource foreign languages widely used in Africa – English, French, and Arabic – simultaneously to provide a single model for cross-lingual transfer learning for African languages. Afro-XLM-R has Afrikaans, Amharic, Hausa, Igbo, Malagasy, Chichewa, Oromo, Nigerian Pidgin, Kinyarwanda, Kirundi, Shona, Somali, Sesotho, Swahili, isiXhosa, Yoruba, and isiZulu. Out of which we have Amharic, Hausa, Igbo, Oromo, Nigerian Pidgin, Kinyarwanda, Swahili and Yoruba are in the shared task. We finetune the base and large models. 

\subsubsection{AfriBERTa}

AfriBERTa is a language model that supports $11$ African languages, including Afaan Oromoo, Amharic, Gahuza (a code-mixed language of Kinyarwanda and Kirundi), Hausa, Igbo, Nigerian Pidgin, Somali, Swahili, Tigrinya, and Yoruba \cite{ogueji-etal-2021-small}. The pretraining corpus for this model is small (only $108.8$ million tokens), when compared to many other language models). AfriBERTa is trained using a Transformer with the standard masked language modelling objective. The AfriBERTa model uses $6$ attention heads, $768$ hidden units, $3072$ feed forward dimensions, and a maximum length of $512$ for the $3$ configurations of the model. We finetune the base and large models. 

% \subsection{AfriTeVa}

% AfriTeVa \cite{jude-ogundepo-etal-2022-afriteva} is an encoder-decoder language model trained on 10 African languages and English using similar training objectives like the T5 model. AfriTeVa has  Afaan Oromoo, Amharic, Gahuza, Hausa, Igbo, Nigerian Pidgin, Somali, Swahili, Tigrinya, Yoruba languages. 

\subsubsection{Serengeti}

Serengeti is an XLM-R based model on pretrained on $517$ African languages, the largest number of African languages in a single model \cite{serengeti_2022}.

\subsubsection{Afro-XLMR\textsubscript{ft}}

Afro-XLMR\textsubscript{ft} is further pretrained using MLM objective on the training data for all tasks. We pretrain for 75 epochs to improve the performance on the sentiment analysis task. 

\subsubsection{Serengeti\textsubscript{ft}}

Serengeti\textsubscript{ft}is further pretrained using MLM objective on the training data for all tasks. We pretrain for 75 epochs to improve the performance on the sentiment analysis task. 
%More details about the training are provided in the following sections. 

\section{Experimental Setup}\label{sec:experiment}

All our models are implemented using the PyTorch framework and the open-source Huggingface Transformers libraries. All the models were trained on a single Nvidia A100. All our models are trained using Adam optimizer with a linear learning rate scheduler. After hyperparameter tuning using Optuna, it was found that the optimum learning rate, batch size and number of epochs is $5*e^-5$, $16$ and $50$ respectively. For the focal loss, the hyper-parameters $\gamma$ and $\alpha$ are set to $2$ and $0.8$, respectively. All models are evaluated on the Weighted $F_1$ Metric which was also used the objective for fine-tuning. 

For further pretraining of Serengeti and Afro-XLMR we used a more aggressive learning rate of $4*e^4$ using a batch size of $16$ for $75$ epochs.

\section{Results}\label{sec:results}
We show the results on the Dev. set for each model in Table \ref{tab:devresults} and results on the Test set in Table \ref{tab:testresults}. The official results from the shared task is labelled as M11 in Table \ref{tab:testresults}. Afro-XLMR-base\textsubscript{ft} (M9) outperforms other models on $5$ languages with an average $F_1$ score of $70.36$ in the Dev. set. Serengeti\textsubscript{ft} (M10) has the second highest performance with an average $F_1$ score of $69.59$ and achieving best performance on $3$ languages on Dev. set. For the Test set, Afro-XLMR-base\textsubscript{ft} (M9) outperforms other models on $9$ languages with an average $F_1$ score of $66.13$ while Serengeti\textsubscript{ft} (M10) has the second highest performance with an average $F_1$ score of $64.97$ and best performance on $1$ language. 
%Please add the following packages if necessary:
%\usepackage{booktabs, multirow} % for borders and merged ranges
%\usepackage{soul}% for underlines
%\usepackage[table]{xcolor} % for cell colors
%\usepackage{changepage,threeparttable} % for wide tables
%If the table is too wide, replace \begin{table}[!htp]...\end{table} with
%\begin{adjustwidth}{-2.5 cm}{-2.5 cm}\centering\begin{threeparttable}[!htb]...\end{threeparttable}\end{adjustwidth}
\begin{table*}[!h]
\centering
\small
\resizebox{\textwidth}{!}{%
\begin{tabular}{lcccccccccccc}\toprule
\textbf{Lang.} &\textbf{M1} & \textbf{M2} &\textbf{M3} &\textbf{M4} &\textbf{M5} &\textbf{M6} &\textbf{M7} &\textbf{M8} &\textbf{M9}&\textbf{M10}&\textbf{M11}&\textbf{Rank} \\\midrule
yo &61.65 &25.33 &65.17 &25.33 &71.02 &72.53 &73.88 &69.63 &\textbf{75.060} &74.82 &71.02 &20th \\
twi &49.58 &30.51 &60.18 &30.51 &63.46 &65.74 &65.24 &46.86 &\textbf{65.950} &65.73 &65.14 &12th \\
ts &35.42 &30.74 &51.05 &30.74 &45.49 &53.07 &49.82 &35.18 &51.62 &\textbf{54.970} &45.49 &28th \\
sw &45.02 &44.22 &51.95 &44.22 &58.60 &\textbf{62.820} &60.58 &60.87 &62.09 &60.40 &58.60 &20th \\
pt &67.37 &51.17 &63.39 &51.17 &65.64 &57.19 &58.37 &61.07 &\textbf{70.670} &61.98 &61.98 &27th \\
pcm &66.24 &40.20 &40.20 &40.20 &67.68 &64.22 &62.99 &61.93 &\textbf{69.500} &65.57 &65.57 &21st \\
ma &52.75 &21.53 &45.14 &21.53 &48.11 &40.60 &45.24 &42.67 &\textbf{59.520} &53.06 &53.06 &22nd \\
kr &53.80 &21.22 &57.27 &21.22 &67.56 &64.12 &62.02 &65.24 &\textbf{69.590} &64.94 &62.02 &23rd \\
ig &75.89 &26.91 &75.79 &26.91 &77.52 &78.41 &79.24 &71.87 &\textbf{79.630} &79.31 &77.52 &17th \\
ha &73.49 &17.02 &73.18 &17.02 &77.60 &79.37 &78.00 &77.30 &\textbf{79.380} &79.37 &79.37 &18th \\
dz &59.30 &32.87 &61.45 &32.87 &64.02 &44.35 &35.96 &37.27 &\textbf{66.570} &60.45 &64.02 &20th \\
am &60.47 &2.26 &2.36 &2.26 &56.88 &\textbf{61.630} &60.62 &53.77 &43.95 &59.02 &56.88 &19th \\
Average &58.42 &28.66 &53.93 &28.66 &63.63 &62.00 &61.00 &56.97 &\textbf{66.13} &64.97 &63.39 & \\
multilingual &61.43 &17.06 &17.06 &17.06 &68.69 &64.84 &65.64 &65.60 &\textbf{69.030} &67.89 &69.03 &- \\
or &36.00 &15.15 &15.15 &15.15 &43.97 &\textbf{50.720} &49.78 &38.20 &44.98 &45.27 &41.79 &14th \\
tg &38.91 &14.38 &14.38 &14.38 &54.38 &40.70 &45.24 &\textbf{57.720} &56.64 &45.73 &57.03 &19th \\
\bottomrule
\end{tabular}}\caption{Results of Model Performance and Rank on Test Set. \textbf{M1:} xlmr-base, \textbf{M2:} xlmr-large, \textbf{M3:} mbert-base-cased, \textbf{M4:} afro-xlmr-large, \textbf{M5:} afro-xlmr-base, \textbf{M6:} afriberta\_large, \textbf{M7:} afriberta\_base, \textbf{M8:} serengeti, \textbf{M9:} afro-xlmr-base\textsubscript{ft}, \textbf{M10:} serengeti\textsubscript{ft}, \textbf{M11:} Official shared task results with Serengeti model}\label{tab:testresults}
\end{table*}

\subsection{Further-Pretraining}
We find significant improvement in model performance after pre-training when compared to fine-tuning. For all but two languages, the further pre-trained LMs - Afro-XLMR-base\textsubscript{ft} (M9) and Serengeti\textsubscript{ft} (M10) outperform their fine-tune counterparts Afro-XLMR-base (M5) and Serengeti (M8). Our findings corroborates research that further pretraining encodes shallow domain knowledge that has influence in low resource scenarios. This is said to be beneficial for providing task specific knowledge for fine-tuning \cite{zhu-etal-2021-pre}.

\subsection{Multi-Lingual Settings}
In multilingual settings, we find that each model achieves $F_1$ scores higher than the average on individual languages. Our finding corroborates research that multilingual training can even achieve better performance than monolingual training, especially for low-resource languages \cite{DBLP:journals/corr/abs-2108-10640}.

\subsection{Zero Shot Settings}
In the zero-shot settings with Oromo and Tigrinya, AfriBERTa-large outperforms other models on Oromo while Serengeti outperforms other models on Tigrinya. In both languages, the further-pretrained models do not achieve best performance. Although further-pretraining improves the performance on Oromo, further-pretraining hurt Serengeti's the performance on Tigrinya. 

\section{Conclusion}\label{sec:conc}
We reported our participation in the three sub-stacks for the AfriSenti-SemEval 2023 shared task. We described our transfer learning approaches using finetuning and further pretraining of existing LMs. We show the performance of our models across the $14$ languages in the three subtask. 

\section{Acknowledgement}
We gratefully acknowledge support from the Natural Sciences and Engineering Research Council of Canada (NSERC; RGPIN-2018-04267), the Social Sciences and Humanities Research Council of Canada (SSHRC; 435-2018-0576; 895-2020-1004; 895-2021-1008), Canadian Foundation for Innovation (CFI; 37771), Compute Canada (CC),\footnote{\href{https://www.computecanada.ca}{https://www.computecanada.ca}} UBC ARC-Sockeye,\footnote{\href{https://arc.ubc.ca/ubc-arc-sockeye}{https://arc.ubc.ca/ubc-arc-sockeye}} and Advanced Micro Devices, Inc. (AMD). Any opinions, conclusions or recommendations expressed in this material are those of the author(s) and do not necessarily reflect the views of NSERC, SSHRC, CFI, CC, AMD, or UBC ARC-Sockeye.

\bibliography{custom}
\bibliographystyle{acl_natbib}

\appendix
\section{Appendix}\label{app:apx}
\subsection{Hausa}

Hausa is a Chadic language spoken by over 50 million people in West Africa. It is tonal, with a diverse vocabulary influenced by Arabic, Fula, and English. Hausa has a long literary tradition, written in a modified Arabic script. It is an important lingua franca and cultural language in West Africa.

\subsection{Yoruba}

Yoruba is a tonal, complex language spoken in Nigeria by over 20 million people. It has a rich vocabulary, oral tradition, and unique script. It conveys meaning through three distinct tones and a noun class system. 

\subsection{Igbo}

Igbo is a tonal language spoken in Nigeria by over 20 million people. It has a rich oral tradition, expressive vocabulary and unique writing system. It conveys meaning through tone variation and has complex sentence structure. %It remains a vital aspect of Igbo culture and identity.

\subsection{Nigerian Pidgin}

Nigerian Pidgin is a creole language that blends English with African languages. It's widely spoken in Nigeria as a lingua franca and has its own unique grammar, vocabulary and pronunciation. 

\subsection{Amharic}

Amharic is a Semitic language spoken in Ethiopia by over 22 million people. It uses the Ethiopian script and is characteristically known for its unique sounds and tonal patterns.

\subsection{Tigrinya}

Tigrinya is a Semitic language spoken in Eritrea and Ethiopia by over 6 million people. It uses a unique script called "Ge'ez" and has a rich oral tradition. Tigrinya is characterized by its distinctive vowel harmonies and use of suffixes.

\subsection{Oromo}

Oromo is a Cushitic language spoken in Ethiopia and Kenya by over 30 million people. It has a unique alphabet called "Qubee" and a rich oral tradition, including folktales and traditional songs. Oromo is characterized by its tonal system and use of suffixes to convey grammatical relationships.

\subsection{Swahili}

Swahili is a Bantu language widely spoken in East Africa, particularly in Kenya and Tanzania. It uses the Latin script and has loanwords from Arabic, Portuguese, and English. Swahili has many variations and dialects, with a rich oral tradition of poetry and song. It is a tonal language, with two distinctive tones that change the meaning of words.

\subsection{Algerian Arabic}

Algerian Arabic is a dialect of Arabic spoken in Algeria. It is characterized by its unique vocabulary, pronunciation, and grammar, as well as the influence of Berber and French. It is written in the Arabic script. 

\subsection{Moroccan Arabic}

Moroccan Arabic, also known as Darija, is a Arabic dialect spoken in Morocco. It has Berber, French, and Spanish influences and uses the Arabic script. Darija is known for its unique pronunciation, vocabulary, and grammar, making it distinct from Standard Arabic.

\subsection{Kinyarwanda}

Kinyarwanda is a Bantu language spoken in Rwanda and Uganda. It uses a unique script called "Kirundi" and has a complex noun class system. It also has a rich oral tradition, with proverbs playing a significant role in the language and culture. Kinyarwanda is characterized by its use of tone to convey meaning and its distinct vowel harmony.

\subsection{Twi}

Twi is a Kwa language spoken in Ghana by over 9 million people. It is tonal and has a rich vocabulary with loanwords from various African and European languages. Twi uses the Latin script and has a long history of oral tradition, including proverbs and folktales.

\subsection{Mozambican Portuguese}

Mozambican Portuguese is a Portuguese dialect spoken in Mozambique. It is characterized by African influences and has evolved differently from European Portuguese. It uses the Latin alphabet and has unique vocabulary and pronunciation. 

\begin{table*}[h!]
\centering
\resizebox{0.9\textwidth}{!}{%
\begin{tabular}{lclc}

\toprule
\textbf{Language}     & \textbf{Code} & \textbf{Classification}                                                                                                                                     & \textbf{Script} \\\midrule
Algerian Arabic       & dz            &  afro-asiatic, semitic, west semitic, central semitic, arabian, & Latin, \\
&&Arabic, north African Arabic, Algerian Arabic                                                                                                                                                            & Arabic          \\ \hdashline
Amharic               & am            & Afro-asiatic, Semitic, South, Ethiopian, South, \\ & & Transversal, Amharic-argobba                                                                               & Ethiopic        \\\hdashline
Hausa                 & ha            & Afro-asiatic, Chadic, west, A, A.1                                                                                                                          & Latin           \\\hdashline
Igbo                  & ig            & Niger-congo, Atlantic congo, volta-congo, benue-congo, \\ & & igboid, igbo                                                                                         & Latin           \\\hdashline
Kinyarwanda           & kr            & \begin{tabular}[c]{@{}c@{}}Niger-congo, Atlantic congo, volta-congo, benue-congo, \\ bantoid, southern, narrow bantu, central, J, Ruanda-rundi\end{tabular} & Latin           \\\hdashline
Moroccan Arabic       & ma            &   afro-asiatic, semitic, west semitic, central semitic, arabian, \\
& & Arabic, north African Arabic, Moroccan-Andalusian Arabic, Moroccan Arabic                                                                                                                                                          & Arabic          \\\hdashline
Mozambican \\
Portuguese & pt            &     Indo-European, classical Indo-European, Italic, Latino-Faliscan, Latinic,\\
&& Imperial Latin, Romance, Italo-Western Romance,  Western Romance, \\
&&Shifted Western Romance, Southwestern Shifted Romance,\\
&& West Ibero-Romance, Galician Romance, Macro-Portuguese, \\
&& Brazil-Portugal Portuguese, Portuguese, 
Nigerian Pidgin
           & Latin           \\\hdashline
Nigerian Pidgin       & pcm           &  Creole-English, English based, Atlantic, Krio                                                                                                                                                           & Latin           \\\hdashline
Oromo                 & or            & Afro-asiatic, Cushitic, East, Oromo                                                                                                                         & Latin           \\\hdashline
Swahili               & sw            & \begin{tabular}[l]{@{}c@{}}Niger-congo, Atlantic congo, volta-congo, benue-congo, \\ bantoid, southern, narrow bantu, central, G, swahili\end{tabular}      & Latin           \\
Tigrinya              & tg            & Afro-asiatic, Semitic, South, Ethiopian, North                                                                                                              & Ethiopic        \\\hdashline
Twi                   & twi           & \begin{tabular}[l]{@{}c@{}}Niger-congo, Atlantic congo, Volta-congo, Kwa, Nyo, \\ Potou-tano, Tano, Central, Akan\end{tabular}                              & Latin           \\\hdashline
Yoruba                & yo            & \begin{tabular}[c]{@{}c@{}}Niger-congo, Atlantic congo, volta-congo, benue-congo, \\ defoid, yoruboid, edekiri\end{tabular}                                 & Latin  \\ \bottomrule        
\end{tabular}%
}
\caption{Details about each language in Afri-Senti Data}
\label{tab:langsinfo}
\end{table*}
% Escape special TeX symbols (%, &, _, #, $)
% Compress whitespace    
%Please add the following packages if necessary:
%\usepackage{booktabs, multirow} % for borders and merged ranges
%\usepackage{soul}% for underlines
%\usepackage[table]{xcolor} % for cell colors
%\usepackage{changepage,threeparttable} % for wide tables
%If the table is too wide, replace \begin{table}[!htp]...\end{table} with
%\begin{adjustwidth}{-2.5 cm}{-2.5 cm}\centering\begin{threeparttable}[!htb]...\end{threeparttable}\end{adjustwidth}
\begin{table*}[!h]
\centering
\small
\resizebox{\textwidth}{!}{%
\begin{tabular}{lccccccccccc}\toprule
\textbf{Lang.} &\textbf{M1} & \textbf{M2} &\textbf{M3} &\textbf{M4} &\textbf{M5} &\textbf{M6} &\textbf{M7} &\textbf{M8} &\textbf{M9}&\textbf{M10} \\\midrule
yo &70.74  &25.14  &71.49  &25.14  &76.43  &78.07  &76.74  &73.34  &78.15  &\textbf{78.72 } \\
twi &47.95  &30.23  &58.55  &30.23  &63.30  &66.80  &62.96  &44.90  &67.72  &\textbf{67.88 } \\
ts &34.80  &30.37  &50.85  &30.37  &48.02  &57.76  &57.25  &34.54  &51.09  &59.33  \\
sw &44.95  &43.98  &51.70  &43.98  &61.33  &\textbf{61.62 } &60.61  &58.27  &59.73  &60.44  \\
pt &68.72  &35.75  &63.71  &35.75  &67.37  &59.04  &58.91  &59.73  &\textbf{70.40 } &66.03  \\
pcm &74.15  &49.28  &49.28  &49.28  &75.90  &72.55  &73.36  &70.79  &\textbf{76.27 } &75.19  \\
ma &\textbf{82.39}  &15.25  &85.29  &15.25  &74.72  &74.13  &64.76  &71.45  &75.92  &75.22  \\
kr &56.05  &21.01  &56.15  &21.01  &68.92  &64.60  &\textbf{81.41 } &67.98  &68.86  &67.01  \\
ig &78.48  &26.94  &78.18  &26.94  &78.92  &80.58  &79.93  &73.97  &80.80  &\textbf{80.82 } \\
ha &76.58  &16.79  &74.84  &16.79  &79.69  &79.59  &43.91  &78.09  &\textbf{81.56 } &79.10  \\
dz &54.97  &37.71  &64.41  &37.71  &65.41  &48.08  &43.91  &45.06  &\textbf{70.75 } &65.94  \\
am &59.81  &35.39  &38.69  &35.39  &62.53  &60.98  &61.44  &59.64  &\textbf{63.05 } &59.38  \\
\midrule
Average &62.46 &30.65 &61.93 &30.65 &68.55 &66.98 &66.71 &61.48 &\textbf{70.36} &69.59 \\
\midrule 
multilingual &68.28  &68.28  &68.28  &68.28  &73.89  &71.67  &71.40  &72.88  &\textbf{75.57 } &73.40  \\ \midrule 
or &36.00  &15.15  &15.15  &15.15  &43.97  &\textbf{50.72 } &49.78  &38.20  &44.98  &45.27  \\
tg &38.91  &14.38  &14.38  &14.38  &54.38  &40.70  &45.24  &\textbf{57.72 } &56.64  &45.73  \\
\bottomrule
\hline
\end{tabular}}\caption{Results of Model Performance on Dev Set. \textbf{M1:} xlmr-base, \textbf{M2:} xlmr-large, \textbf{M3:} mbert-base-cased, \textbf{M4:} afro-xlmr-large, \textbf{M5:} afro-xlmr-base, \textbf{M6:} afriberta\_large, \textbf{M7:} afriberta\_base, \textbf{M8:} serengeti, \textbf{M9:} afro-xlmr-base\textsubscript{ft}, \textbf{M10:} serengeti\textsubscript{ft}.}\label{tab:devresults}
\end{table*}
% \label{sec:appendix}

% This is a section in the appendix.

\end{document}